\documentclass[letterpaper]{article} 
\usepackage{aaai2026}  
\usepackage{times}  
\usepackage{helvet}  
\usepackage{courier}  
\usepackage[hyphens]{url}  
\usepackage{graphicx} 
\usepackage{natbib}  
\usepackage{caption} 
\usepackage{algorithm}
\usepackage{algorithmic}
\usepackage{booktabs}
\usepackage{multirow}
\usepackage{subfigure}
\usepackage{bm}
\usepackage{pdfpages}
\usepackage{newfloat}
\usepackage{listings}
\DeclareCaptionStyle{ruled}{labelfont=normalfont,labelsep=colon,strut=off} 
\floatstyle{ruled}
\newfloat{listing}{tb}{lst}{}
\floatname{listing}{Listing}
\title{IndoorUAV: Benchmarking Vision-Language UAV Navigation in Continuous Indoor Environments}
\author{
    Xu Liu\textsuperscript{\rm 1}\equalcontrib, Yu Liu\textsuperscript{\rm 1}\equalcontrib, Hanshuo Qiu\textsuperscript{\rm 1}, Yang Qirong\textsuperscript{\rm 1}, Zhouhui Lian\textsuperscript{\rm 1, 2}\thanks{Corresponding author}
}
\affiliations{
    \textsuperscript{\rm 1}Wangxuan Institute of Computer Technology, Peking University, Beijing, China\\
    \textsuperscript{\rm 2}State Key Laboratory of General Artificial Intelligence, Peking University, Beijing, China\\

    2301213301@stu.pku.edu.cn, yuliu\_@hust.edu.cn, qiuhanshuo@nudt.edu.cn, \\
    2400013185@stu.pku.edu.cn, lianzhouhui@pku.edu.cn
}

\usepackage{bibentry}
\usepackage{amsmath}

\begin{document}

\maketitle

\begin{abstract}
Vision-Language Navigation (VLN) enables agents to navigate in complex environments by following natural language instructions grounded in visual observations. Although most existing work has focused on ground-based robots or outdoor Unmanned Aerial Vehicles (UAVs), indoor UAV-based VLN remains underexplored, despite its relevance to real-world applications such as inspection, delivery, and search-and-rescue in confined spaces. 
To bridge this gap, we introduce \textbf{IndoorUAV}, a novel benchmark and method specifically tailored for VLN with indoor UAVs. We begin by curating over 1,000 diverse and structurally rich 3D indoor scenes from the Habitat simulator. Within these environments, we simulate realistic UAV flight dynamics to collect diverse 3D navigation trajectories manually, further enriched through data augmentation techniques. Furthermore, we design an automated annotation pipeline to generate natural language instructions of varying granularity for each trajectory. This process yields over 16,000 high-quality trajectories, comprising the \textbf{IndoorUAV-VLN} subset, which focuses on long-horizon VLN. 
To support short-horizon planning, we segment long trajectories into sub-trajectories by selecting semantically salient keyframes and regenerating concise instructions, forming the \textbf{IndoorUAV-VLA} subset. 
Finally, we introduce \textbf{IndoorUAV-Agent}, a novel navigation model designed for our benchmark, leveraging task decomposition and multimodal reasoning.
We hope IndoorUAV serves as a valuable resource to advance research on vision-language embodied AI in the indoor aerial navigation domain.
\end{abstract}

\begin{links}
    \link{Datasets}{https://www.modelscope.cn/datasets/valyentine/Indoor_UAV}
\end{links}

\begin{figure}[t]
\centering
\includegraphics[width=\columnwidth]{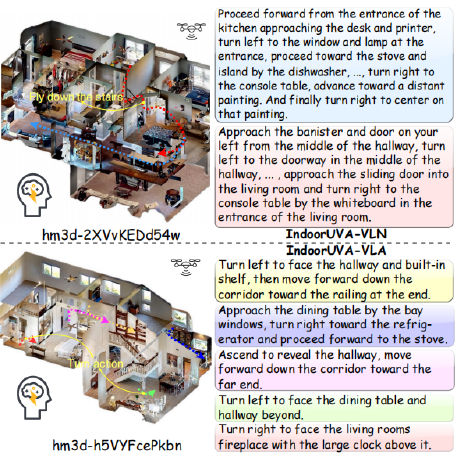}
\caption{Illustration of IndoorUAV-VLN (upper) and IndoorUAV-VLA (lower) datasets. Long-horizon VLN tasks typically involve complex instructions and longer trajectory lengths, while VLA tasks focus on fine-grained maneuver execution, consisting of 1-3 executable actions.}
\label{fig1}
\end{figure}
\section{Introduction}


Vision-Language Navigation (VLN) is a foundational task in embodied AI that aims to enable autonomous agents to follow natural language instructions and navigate complex environments. During the past few years, a wide range of benchmarks and models have been developed for VLN tasks, primarily in two main settings: (1) indoor environments with ground-based agents such as wheeled robots or quadrupeds \cite{anderson2018vision,krantz2020beyond,han2025roomtour3d,zhang2024uni}, and (2) outdoor environments with aerial platforms such as Unmanned Aerial Vehicles (UAVs) \cite{fan2022aerial,liu2023aerialvln,wang2025uav}. These two lines of work have significantly advanced the capabilities of VLN. However, they overlook a critical and practically relevant domain: UAV-based vision-language navigation in complex, cluttered, and structured 3D continuous indoor environments.

This new setting introduces a unique set of challenges that diverge significantly from both ground-based indoor VLN and outdoor aerial VLN. In indoor environments, ground agents are typically confined to 2D navigation on a planar surface, limiting their ability to reason about or interact with the full 3D structure of the space. As a result, widely-used benchmarks such as R2R \cite{anderson2018vision}, SOON \cite{zhu2021soon}, and RxR \cite{ku2020room} are inherently ill-suited for aerial navigation tasks that demand vertical reasoning, free-form 3D maneuvering, and fine-grained spatial understanding. 
Although recent work has started to explore UAV-based VLN in outdoor environments, these scenarios are typically open and sparse, lacking dense obstacles, narrow corridors, and high-precision maneuvering requirements that define indoor spaces, making it difficult to transfer such models to generalize such models to real-world indoor applications.

To bridge this gap, we introduce IndoorUAV benchmark specifically designed for aerial VLN in 3D indoor environments. Our goal is to facilitate research on navigation agents that must interpret natural language instructions and execute flight paths in fully three-dimensional, cluttered indoor spaces with realistic constraints. Unlike previous work, IndoorUAV emphasizes both high-level instruction grounding and low-level aerial control in a unified setting. It serves as a testbed for developing embodied agents capable of complex spatial reasoning, obstacle avoidance, and motion planning—capabilities that are critical for real-world applications such as indoor inspection, search and rescue, and autonomous delivery.

To construct IndoorUAV, we first curate over 1,000 high-quality 3D indoor environments from the Habitat simulator \cite{szot2021habitat,habitat19iccv}. Within these environments, we simulate plausible UAV flight dynamics to collect rich 3D trajectory data manually. To increase data diversity, we apply a variety of trajectory augmentation strategies, including trajectory reverse and sub-trajectory recombination. Then an automatic annotation pipeline is developed to generate natural language instructions at two levels of granularity. This process yields IndoorUAV-VLN, a long-horizon navigation dataset containing over 16,000 instruction-trajectory pairs. This dataset challenges agents to interpret complex, multi-step language instructions and navigate accordingly in 3D space.

Furthermore, to promote fine-grained spatial understanding and short-term planning navigation, we derive IndoorUAV-VLA from IndoorUAV-VLN by segmenting long trajectories into short sub-trajectories via keyframe selection. We then regenerate concise instructions corresponding each sub-trajectory. Each instruction in IndoorUAV-VLA maps to only 1 to 3 UAV actions, offering a complementary benchmark that emphasizes low-level flight control.


Finally, we propose IndoorUAV-Agent, leveraging a LLM to decompose long-horizon natural language instructions into shorter sub-instructions, which are then sequentially executed by a VLA model. This hierarchical design enables the agent to handle complex multi-step tasks while maintaining fine-grained control over UAV motion. 

Experimental results highlight the significant challenges posed by IndoorUAV, revealing a substantial performance gap between current state-of-the-art models and the demands of real-world indoor UAV navigation. These findings underscore the need for further research on grounded language understanding, 3D spatial reasoning, and fine-grained motion control in aerial settings.

In summary, our contributions are as follows:

\begin{itemize}
\item We introduce IndoorUAV, the first large-scale benchmark specifically targeting UAV-based VLN in 3D indoor environments. 
\item We develop an automated data collection and annotation pipeline that generates realistic UAV flight trajectories and multi-granularity natural language instructions.
\item We propose IndoorUAV-Agent, a strong baseline model tailored to the unique challenges of indoor aerial VLN across different navigation settings.
\end{itemize}

\begin{figure*}[t]
\centering
\includegraphics[width=\textwidth]{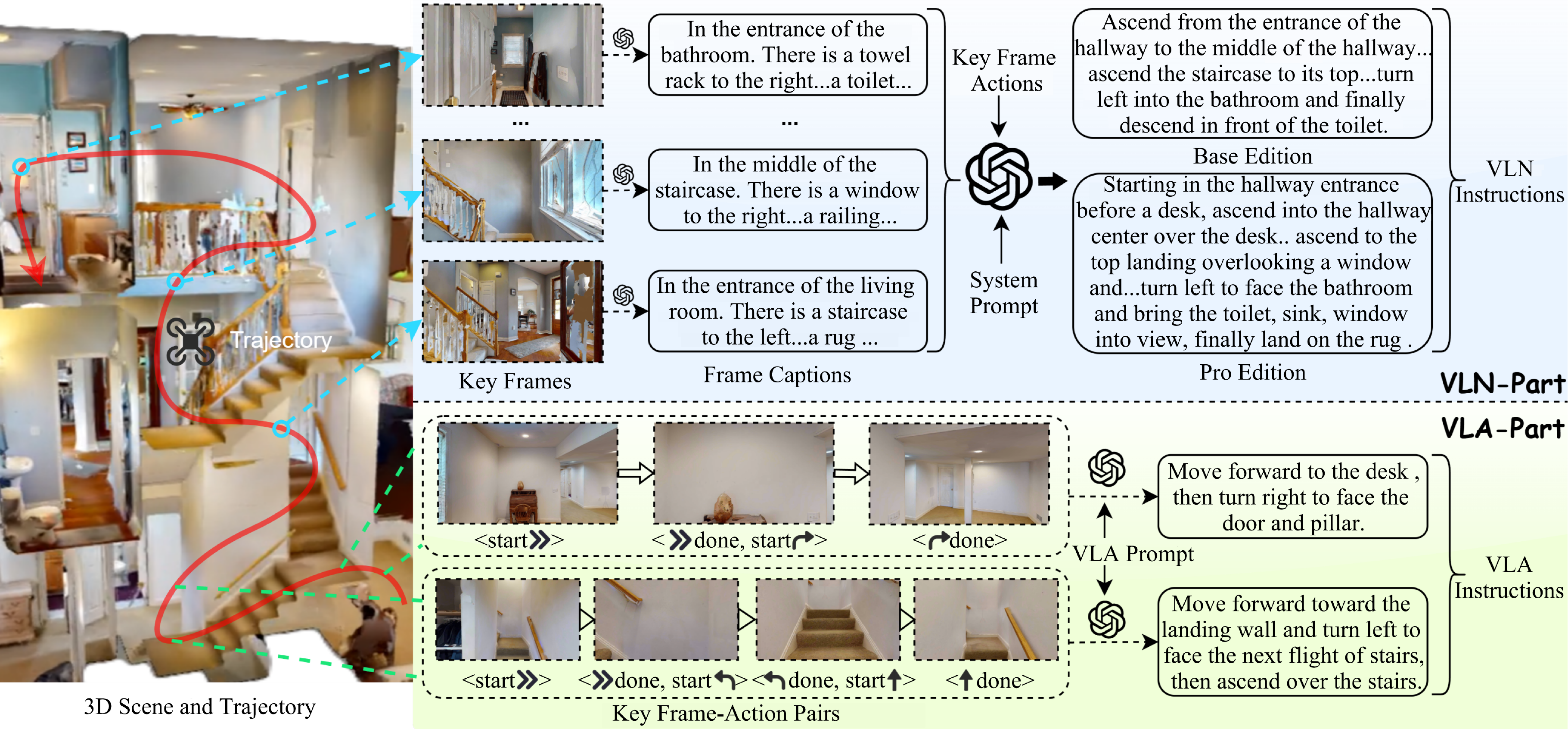}
\caption{Overview of the IndoorUAV data collection and instruction generation pipeline.}
\label{pipe}
\end{figure*}

\section{Related Work}
\subsection{Ground-based Vision-Language Navigation Datasets}
Early VLN datasets focus on ground-based navigation. The upper part of Table \ref{tab:vln_datasets} presents recent ground-based datasets. Room-to-Room (R2R) \cite{anderson2018vision} is the first VLN dataset to enable navigation in discrete environments by using panoramic images. Follow-up datasets expanded this setting. For example, Room-Across-Room (RxR) \cite{ku2020room} extended R2R to multiple languages and longer trajectories. TouchDown \cite{chen2019touchdown} leverages Google Street View to address the challenges of outdoor long-range navigation. CVDN \cite{thomason2020vision} introduces the navigation from dialog history task. All the above datasets are graph-based, with predefined navigable points. VLN-CE \cite{krantz2020beyond} lifts the VLN task to continuous 3D environments by removing the nav-graph. Recent work like LHPR-VLN \cite{song2025towards} emphasizes long-term planning and decision consistency across consecutive subtasks, while OctoNav \cite{gao2025octonav} focuses on generalist embodied navigation.

\subsection{Aerial Vision-Language Navigation Datasets}
Aerial VLN has recently been gaining significant momentum. Unlike ground-based navigation, Aerial VLN involves 3D trajectories, typically has more degrees of freedom, and primarily operates in outdoor environments. The lower part of Table \ref{tab:vln_datasets} presents the primary Aerial VLN datasets. AVDN \cite{fan2022aerial} comprises ~3K continuous drone trajectories in a photorealistic simulator paired with asynchronous human–human dialogs, supporting dialog-guided aerial navigation tasks. OpenFly \cite{gao2025openfly} delivers a comprehensive aerial VLN platform with 100K automatically generated trajectories across 18 diverse scenes, combining multiple rendering engines. UAV‑Flow \cite{wang2025uav} is the first real‑world benchmark for fine-grained, language‑conditioned UAV control, with large-scale real and simulated datasets focusing on short‑horizon navigation.

\subsection{Vision-Language Navigation Methods}
Early VLN agents typically employ LSTM \cite{graves2012long} to process language instructions and visual observations, such as Seq2Seq \cite{anderson2018vision} and CMA \cite{krantz2020beyond}. To more effectively encode historical navigation information, graph-based methods \cite{zhu2021soon,deng2020evolving,wang2021structured,chen2022think} build structured memory in the form of topological or semantic maps. Recent work explore leveraging LLM as planners. NavGPTv2 \cite{zhou2024navgpt} incorporates a Vision Language Model (VLM) and a graph-based policy network for effective action prediction. Uni-NaVid \cite{zhang2024uni} leverages a video-based VLM to unify different paradigms of navigation tasks and propose an online token merge strategy to efficiently process video streams. NavAgent \cite{liu2024navagent} undertakes UAV navigation tasks by synthesizing multi-scale environmental information. SkyVLN \cite{li2025skyvln} incorporates an NMPC module for dynamic obstacle avoidance during the process of UAV navigation.

\begin{table*}[ht]
\centering
\small
\resizebox{\textwidth}{!}{
\begin{tabular}{lcccccccc}
\toprule
\textbf{Dataset} & $N_{\text{traj}}$ & $N_{\text{vocab}}$ & Path Len. & Intr Len. & Action Space & $N_{\text{scenes}}$ & Environment \\
\midrule
R2R~\cite{anderson2018vision} & 7189 & 3.1K & 10.0 & 29  & graph-based  & 90 & Matterport3D \\
RxR~\cite{ku2020room} & 13992 & 7.0K & 14.9 & 129 & graph-based  & 90 & Matterport3D \\
CVDN~\cite{thomason2020vision} & 7415 & 4.4K & 25.0 & 34  & graph-based  & 83 & Matterport3D \\
TouchDown~\cite{chen2019touchdown} & 9326 & 5.0K & 313.9 & 90 & graph-based  & - & Google Street View \\
VLN-CE~\cite{krantz2020beyond} & 4475 & 4.3K & 11.1 & 19  & 2 DoF & 90 & Matterport3D \\
LHPR-VLN~\cite{song2025towards} & 3260 & 0.5K & - & 18.17  & 2 DoF  &  216 & Habitat \\
OctoNav~\cite{gao2025octonav} & 45k & - & - & -  & 2 DoF   & 438  & Mp3D, Gibson, HM3D, ProcTHOR\\
\midrule
AVDN~\cite{fan2022aerial} & 6269 & 3.3K & 144.7 & 89  & 3 DoF & - & xView \\
AerialVLN~\cite{liu2023aerialvln} & 8446 & 4.5K & 661.8 & 83  & 4 DoF  & 25 & AirSim + UE \\
CityNav~\cite{lee2024citynav} & 32637 & 6.6K & 545   & 26  & 4 DoF   & 34 & SensatUrban \\
OpenUAV~\cite{wang2024towards} & 12149 & 10.8K & 255   & 104 & 6 DoF  & 22 & AirSim + UE \\
OpenFly~\cite{gao2025openfly} & 100K & 15.6K & 99.1  & 59  & 4 DoF  & 18 & AirSim, GTA5, 3D GS, GE \\
UAVFlow~\cite{wang2025uav} & 40801 & 0.3K & 10.7 & 9 & 6 DoF  & -  & Real world, UnrealCV\\
\midrule
IndoorUAV-VLN & 16040 & 3.9K & 21.6 & 112 & 4 DoF  & 1075 & Mp3D, Gibson, HM3D, Replica\\
IndoorUAV-VLA &  34925 & 2.2K & 2.2 & 14.5 & 4 DoF  & 1075 & Mp3D, Gibson, HM3D, Replica\\
\bottomrule
\end{tabular}
}
\caption{Comparison between existing VLN (Vision-and-Language Navigation) datasets and our IndoorUAV. Above the middle dividing line lies the ground-based datasets, while below is the aerial VLN datasets. $N_{\text{traj}}$: the number of total trajectories. $N_{\text{vocab}}$: vocabulary size. Path Len: the average length of trajectories, measured in meters. Intr Len: the average length of instructions. $N_{\text{scenes}}$: the number of used scenes.}
\label{tab:vln_datasets}
\end{table*}

\section{IndoorUAV Benchmark}
To advance research in vision-language navigation for indoor aerial agents, we introduce the IndoorUAV benchmark—a large-scale, high-quality dataset specifically designed for simulating realistic UAV-based navigation in cluttered 3D indoor environments. IndoorUAV aims to address the unique challenges of indoor aerial VLN, including unconstrained 3D movement, fine-grained spatial understanding, and real-world-style language instruction. IndoorUAV benchmark consists of over 50000 high-resolution UAV trajectories paired with richly annotated natural language instructions. The benchmark contains two subsets: IndoorUAV-VLN, which focuses on long-horizon, goal-directed navigation through multi-step natural language commands; IndoorUAV-VLA, which targets short-horizon, low-level action planning.
This section details the environment sources, data collection methodology, and the composition of the dataset.

\subsection{Environment Source}
We construct the IndoorUAV benchmark within high-fidelity simulated 3D indoor environments drawn from four widely-used 3D indoor space datasets in the embodied AI community: Matterport3D (MP3D), Gibson, HM3D, and Replica. These environments encompass a diverse range of residential, office, and public spaces, offering rich variability in layout, appearance, and object distribution. From these sources, we manually select and curate over 1,000 high-quality scenes that support rich 3D exploration and navigation. All selected environments are compatible with the Habitat simulator, enabling efficient data generation with realistic physics and photorealistic rendering.

\subsection{Data Collection}
\begin{figure}[t]
\centering
    \subfigure[]{
        \includegraphics[width=\columnwidth]{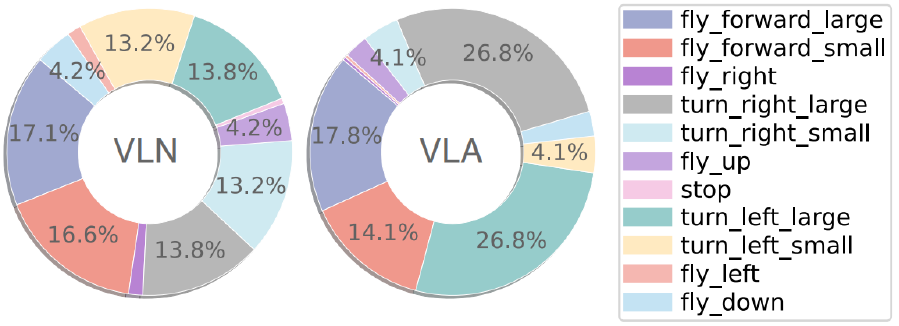}
        \label{fig:2a}
    }
    \subfigure[]{
        \includegraphics[width=\columnwidth]{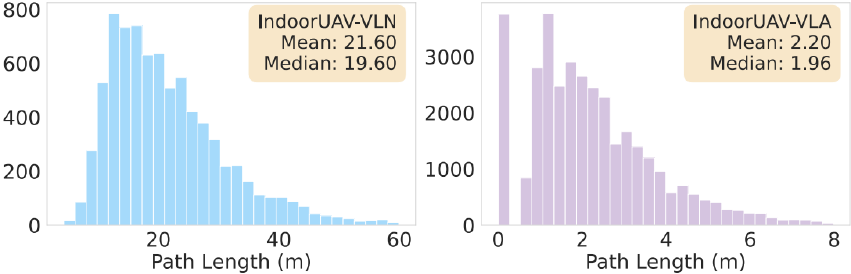}
        \label{fig:2b}
    }
\caption{Statistical analysis of the IndoorUAV benchmark. (a) Action distributions. (b) trajectory length distributions. }
\label{fig:data_analysis}
\end{figure}

To enable unrestricted 3D UAV movement, we first remove the built-in navigation mesh (navmesh) constraints from the environmints, allowing agents to move freely in three dimensions of space. We define a 4 degree-of-freedom (4 DoF) UAV action space, consisting of (1) forward movement, (2) vertical translation (up/down), (3) lateral motion (left/right), and (4) yaw rotation (see Figure \ref{fig:2a}). This configuration closely mirrors realistic UAV flight dynamics and supports the generation of expressive and complex flight trajectories.

\subsubsection{IndoorUAV-VLN Collection}
For the Vision-and-Language Navigation (VLN) subset of IndoorUAV, we first sample a large number of start and goal locations within each scene to define diverse navigation tasks covering various spatial scales and complexity levels. Experienced operators then remotely control the UAV within the simulator to manually navigate from predefined start points to target goals. This manual piloting captures natural, smooth flight paths that reflect realistic human UAV operation in complex indoor settings. 
To augment the dataset without sacrificing quality, we reverse the direction of each trajectory (by swapping start and goal points) and adjust the orientation sequence accordingly. This simple yet effective technique doubles the dataset size while maintaining trajectory realism.

To generate language instructions for each trajectory, we employ a multi-step pipeline built on top of a GPT-4-based instruction generation module (Figure \ref{pipe}). For each trajectory, we extract keyframes based on significant motion changes, such as sharp turns (over 45°), vertical climbs (over 1m), or long linear flights. For each keyframe, we label its corresponding action type (e.g., \text{\texttt{fly\_up}}, \text{\texttt{turn\_right}}) based on the preceding motion segment. Next, we prompt GPT-4o to describe each keyframe image by: (1) describing the location (e.g., kitchen); (2) describing nearby objects and structures using a coarse-to-fine scale (near/mid/far and left/center/right). These image-grounded descriptions are concatenated and passed to GPT-4 again to generate the full navigation instruction. Finally, each trajectory is paired with two levels of instruction: (1) detailed, long-form instruction capturing all spatial and semantic nuances; (2) relatively shorter instruction focusing only on coarse goal descriptions.
IndoorUAV-VLN focuses on long-horizon navigation, requiring the agent to understand and execute multi-step, semantically rich commands in cluttered 3D space.

\subsubsection{IndoorUAV-VLA Collection}
To enable research on low-level control and fine-grained action prediction, we construct a complementary dataset derived from IndoorUAV-VLN. In this subset, we segment each long trajectory into multiple short sub-trajectories, each covering only 1–3 key actions (e.g., “fly forward past the cabinet,” “descend near the table”). For each segment, we regenerate concise navigation instructions that focus on local spatial goals and immediate surroundings by directly prompt the GPT-4o with images. As shown in the Table \ref{tab:vln_datasets}, IndoorUAV-VLA contains 34,925 short trajectories, but the average instruction length is only 14.5 words.

\begin{figure*}[t]
\centering
\includegraphics[width=0.97\textwidth]{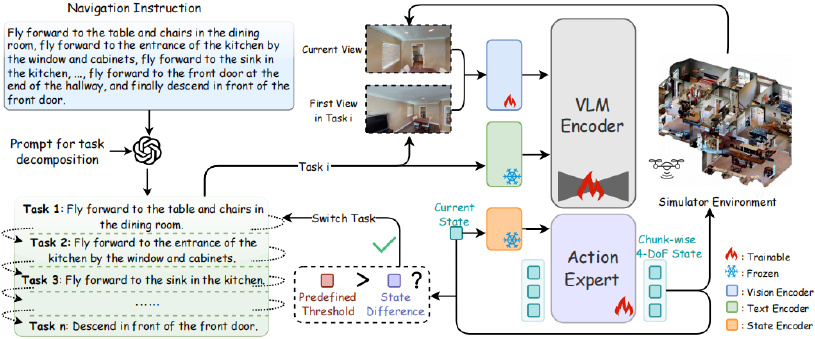}
\caption{For the long-horizon VLN task, we first use GPT-4o to decompose the long instruction into $n$ shorter VLA-style instructions as subtasks, each containing 1 to 3 actions. We then process each subtask sequentially using a VLA model based on the \(\bm{\pi_{0}}\) architecture.}
\label{model_pipeline}
\end{figure*}

\subsection{Dataset Analysis}
In this section, we conduct a comprehensive analysis of IndoorUAV from multiple perspectives, including scene diversity, trajectory complexity, and instruction granularity.

As shown in Table \ref{tab:vln_datasets}, IndoorUAV is constructed from a diverse collection of 1,075 photorealistic indoor scenes, significantly surpassing all previous VLN datasets in terms of scene diversity. In total, the dataset comprises 50965 navigation trajectories, each representing a 4-DoF UAV trajectory that supports horizontal translation, vertical movement, and yaw rotation.
Figure~\ref{fig:2a} presents the overall distribution of discrete actions across the two dataset parts. The most frequent actions are \text{\texttt{fly\_forward}}, \text{\texttt{turn\_left}} and \text{\texttt{turn\_right}}, which are essential for navigating tight indoor spaces. To improve balance across action types and enable finer control, we define dual-scale versions of key movements: \text{\texttt{fly\_forward\_small}} moves the UAV forward by 0.15 meters, while \texttt{fly\_forward\_large} performs a 0.9 meters step; \texttt{turn\_left\_small} and \texttt{turn\_right\_small} rotate the UAV by \textbf{3°}, while \texttt{turn\_left\_large} and \texttt{turn\_right\_large} perform turns of \textbf{15°}. This fine-grained action space is particularly suited for indoor aerial navigation, where small adjustments are critical for avoiding obstacles and maintaining stable flight.

Figure~\ref{fig:2b} illustrates the distribution of trajectory lengths across the two subsets of IndoorUAV. To provide a more systematic characterization of task complexity, we annotate each trajectory with a difficulty level. Specifically, in IndoorUAV-VLN, trajectories containing fewer than 120 actions are categorized as easy, those with 120–200 actions as medium, and those exceeding 200 actions as hard. For IndoorUAV-VLA, difficulty levels are defined based on action types, with 1 type denoting easy, 2 types denoting medium, and 3 types denoting hard.


\section{IndoorUAV-Agent}
\subsection{Task Definition}
Formally, an instruction $I$ describes a navigation goal composed of multiple semantic steps (e.g., "Exit the kitchen, fly down the hallway, and enter the second room on the right."). The agent starts at an initial pose $s_0 = (x_0, y_0, z_0, \theta_0)$ and must predict a sequence of poses that lead to the successful completion of the task. We divide the overall problem into two categories: (1) Short-horizon VLA tasks, where the instruction corresponds to 1–3 primitive actions and can be directly translated into a continuous low-level trajectory. (2) Long-horizon VLN tasks, where the instruction is compositional and must be decomposed into a sequence of simpler navigation goals.

\subsection{Model Architecture}
For short-horizon VLA tasks, we directly leverage a fine-tuned \(\bm{\pi_{0}}\) model to predict a horizon of $h$ future robot states, including the 3D coordinates $(x, y, z)$ and yaw angle $\theta$ at each step. This model takes as input the current egocentric visual observation and a short natural language instruction, and outputs continuous low-level controls in the form of a predicted trajectory. The \(\bm{\pi_{0}}\) architecture, with its language-conditioned visual encoder and trajectory decoder, proves effective for such short and precise navigation tasks that typically involve only one or two atomic actions.

\begin{table*}[ht]
\centering
\begin{tabular}{c|cc|cc|cc|cc}
\toprule
\multirow{2}{*}{\textbf{IndoorUAV-VLA}} & 
\multicolumn{2}{c|}{\textbf{Full}} & 
\multicolumn{2}{c|}{\textbf{Easy}} & 
\multicolumn{2}{c|}{\textbf{Medium}} & 
\multicolumn{2}{c}{\textbf{Hard}} \\
& 
SR/\%$\uparrow$ & NDTW/\%$\uparrow$ & SR/\%$\uparrow$ & NDTW/\%$\uparrow$ & 
SR/\%$\uparrow$ & NDTW/\%$\uparrow$ & SR/\%$\uparrow$ & NDTW/\%$\uparrow$ \\
\midrule
GPT-4o & 11.69 & 9.2 & 30.30 & 12.30 & 4.00 & 6.57 & 1.96 & 4.84 \\
Seq2Seq$^{\ast}$ & 1.33 & 2.74 & 1.6 & 2.63 & 1.2 & 2.74 & 1.03 & 3.03\\
CMA$^{\ast}$ & 0.99 & 1.88 & 1.28 & 1.75 & 0.75 & 1.94 & 1.03 & 2.01\\
OpenVLA$^{\ast}$ & 7.81 & 2.42 & 22.52 & 2.89 & 1.19 & 1.12 & 0.0 & 0.12\\
\(\bm{\pi_{0}}\)-FAST$^{\ast}$ & 8.62 & 4.71 & 18.09 & 8.83 & 5.26 & 2.93 & 1.14 & 2.68\\
NaVid$^{\ast}$  & 15.82 & 5.28 & 25.31 & 13.1 & 18.21 & 3.21 & 2.31 & 1.72 \\
\(\bm{\pi_{0}}\)$^{\ast}$ & 27.16 & 9.44 & 46.58 & 14.52 & 21.64 & 7.64 & 7.55 & 4.27\\
\bottomrule
\end{tabular}
\caption{Comparison results on the IndoorUAV-VLA test split. $^{\ast}$ indicates models fine-tuned on the dataset.}
\label{tab:vla_results}
\end{table*}

\begin{table*}[ht]
\centering
\begin{tabular}{c|cccc|cccc}
\toprule
\multirow{2}{*}{\textbf{IndoorUAV-VLN}} & 
\multicolumn{4}{c|}{\textbf{TEST SEEN}} & 
\multicolumn{4}{c}{\textbf{TEST UNSEEN}} \\
& 
NE/m$\downarrow$ & SR/\%$\uparrow$ & OSR/\%$\uparrow$ & NDTW/\%$\uparrow$ & 
NE/m$\downarrow$ & SR/\%$\uparrow$ & OSR/\%$\uparrow$ & NDTW/\%$\uparrow$ \\
\midrule
GPT-4o & 7.96 & 3.69 &4.67  & 6.96 & 8.52 & 0.56 & 2.78  & 6.82 \\
Seq2Seq$^{\ast}$ & 10.18 & 0.89 & 2.23 & 1.41 & 11.6 & 0.41 & 1.85 & 1.23\\
CMA$^{\ast}$ & 11.6 & 1.56 & 8.48 & 1.39 & 12.15 & 1.64 & 10.47 & 0.81\\
\(\bm{\pi_{0}}\)$^{\ast}$ & 8.35 & 2.92 & 11.5 & 11.87 & 8.81 & 2.83 & 10.02 & 11.69 \\
NaVid$^{\ast}$ & 21.83 & 0.75 & 14.70 & 1.36 & 19.40 & 0.84& 16.21 &  2.32 \\
OpenFly-Agent$^{\ast}$ & 8.17 & 4.12 & 10.96 & 10.63 & 8.83 & 2.58 & 9.45& 10.14 \\
Ours & 6.62 & 7.29 & 12.83 & 17.19 & 7.27 & 5.06 & 13.49 & 15.65\\

\bottomrule
\end{tabular}
\caption{Comparison results on the IndoorUAV-VLN test split. $^{\ast}$ indicates models fine-tuned on the dataset.}
\label{tab:vla_results}
\end{table*}

In contrast, long-horizon VLN tasks often involve complex instructions spanning multiple semantic steps (e.g., "Exit the room, go down the hallway, and turn left at the second door"), where current end-to-end continuous control models struggle to generalize. To address this, we propose a task decomposition pipeline (see Figure \ref{model_pipeline}), which first utilizes GPT-4o to split a long-horizon instruction into $n$ shorter, VLA-style sub-instructions, each describing a simple goal that can typically be completed in 1–3 actions. These sub-instructions are then processed sequentially, where for each subtask we invoke the \(\bm{\pi_{0}}\)-based VLA model to predict and execute the corresponding trajectory segment.
Importantly, for the $i$-th subtask, we use the predicted final-state observation (image) from the $(i{-}1)$-th subtask as the first reference frame for the current model input. This design ensures temporal continuity across subtasks and mitigates error accumulation by aligning each instruction with an updated visual context.
By interleaving natural language understanding and low-level motion planning in this modular fashion, our approach improves interpretability, robustness, and success rate on complex VLN tasks.

\section{Experiments}

\subsection{Evaluation Metrics}
Following the previous work, we utilize four metrics for evaluation. (1) \textbf{Success Rate (SR)}: The definition of SR varies depending on the task type.
For \textit{VLA} tasks, a trajectory is considered successful if the predicted final position is within 0.5 meters of the target and the yaw difference is less than \(\frac{\pi}{4}\).
For \textit{VLN} tasks, where high-level navigation goals are given, a prediction is considered successful if the final position is within 2 meters of the target.
(2) \textbf{Normalized Dynamic Time Warping (NDTW)} \cite{ilharco2019general}: While some predicted trajectories may be semantically correct, they may follow irregular or suboptimal paths. To account for such discrepancies, we additionally compute NDTW to quantitatively assess the alignment between predicted and reference trajectories. For VLA tasks, we calculate both the three-dimensional coordinate NDTW and yaw angle NDTW for each trajectory, and weighted them respectively based on the path length and cumulative rotation angle of the trajectory as the final NDTW. And for VLN tasks, we only calculate the three-dimensional coordinate NDTW. (3) \textbf{Navigation Error (NE)}: This metric measures the distance between the final position of the predicted trajectory and the target position. (4) \textbf{Oracle Success Rate (OSR)}: OSR evaluates whether any point along the predicted trajectory satisfies the success condition defined in SR.

\subsection{Baseline Models}
We benchmark various representative models spanning VLA and VLN. The VLA models directly output continuous robot motion controls informed by visual observations and language instructions. In contrast, VLN models predict high‑level navigation steps from visual inputs and language, which are then executed as motion. 
\subsubsection{VLA Models}
(1) \(\bm{\pi_{0}}\) \cite{black2410pi0} and \(\bm{\pi_{0}}\)-FAST \cite{pertsch2025fast}: build on a pretrained VLM and use diffusion-style decoding to generate smooth continuous control sequences at up to 50 Hz. (2) \textbf{OpenVLA} \cite{kim2024openvla}: An open-source discrete-token VLA model trained on ~970K real‑robot episodes, which achieves broad generalization and strong zero‑shot and fine‑tuned performance.

At each time step \(t\), the agent receives the initial observation \(o_1\), current observation \(o_t\), and current state \(s_t\). The state is defined as a tuple of 3D coordinates and a yaw angle, i.e., \(s_t = (x_t, y_t, z_t, \theta_t)\). The VLA model encodes these inputs to predict a sequence of future states for the next \(h\) time steps:
\begin{equation}
\begin{split}
S_{t+1:t+h+1} = 
\text{Model}_{\text{VLA}}(\mathcal{O}_{1}, \mathcal{O}_{t}, \mathcal{I}, s_t)
\end{split}
\end{equation}

\begin{figure*}[t]
\centering
\includegraphics[width=\textwidth]{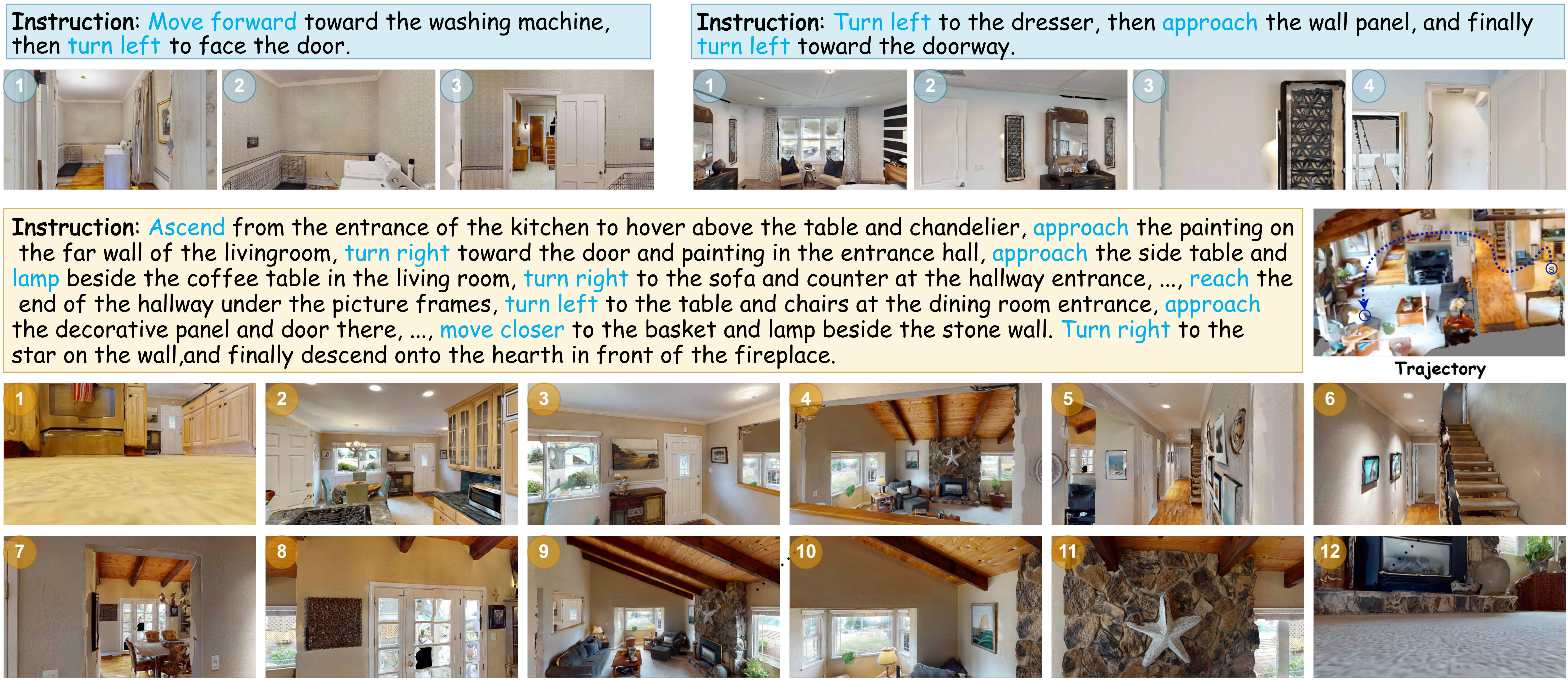}
\caption{Visualization on both IndoorUAV-VLA and IndoorUAV-VLN. The upper two are VLA tasks, with medium/hard (2-3 executable actions) difficulty, respectively. The lower is a VLN task where the markers \textit{S} and \textit{T} in the trajectory plot indicate the start position and target position of the trajectory, respectively.}
\label{vis}
\end{figure*}

\subsubsection{VLN Models}
(1) \textbf{Seq2Seq} and \textbf{CMA} \cite{krantz2020beyond}: Traditional LSTM-style models that encode visual inputs and instructions to regress continuous waypoint sequences using a recurrent policy. (2) \textbf{NaVid} \cite{zhang2024navid}: A video-based VLM trained with a large amount of data , which is designed for ground-based VLN. 
(3) \textbf{OpenFly-Agent} \cite{gao2025openfly}: an aerial navigation model that builds upon the OpenVLA.

At each time step \(t\), the agent receives a navigation instruction \(\mathcal{I}\), along with a sequence of past observations \(\mathcal{O}_{1:t} = \{o_1, o_2, \dots, o_t\}\). The VLN model encodes both modalities and produces a navigation action \(a_t \in \mathcal{A}\), where \(\mathcal{A}\) denotes a discrete action space. Formally:
\begin{equation}
\begin{split}
a_t = \text{Model}_{\text{VLN}}(\mathcal{O}_{1:t}, \mathcal{I})
\end{split}
\end{equation}

\subsection{Quantitative results}
\paragraph{IndoorUAV-VLA.}
Table ~\ref{tab:vla_results} summarizes the performance of various models on the IndoorUAV-VLA task. Among all methods, fine-tuned \(\bm{\pi_{0}}\) achieves the best overall results, attaining 27.16\% Success Rate (SR) and 9.44\% NDTW across the full test split. On easy trajectories, it achieves an impressive 46.58\% SR, significantly outperforming other methods, and remains competitive on medium (21.64\%) and hard (7.55\%) tasks. Traditional VLN-style regression models fail to generalize to the low-level action space, with SR below 3\%. NaVid shows improved performance but still falls short of the results achieved by \(\bm{\pi_{0}}\).

\paragraph{IndoorUAV-VLN.}
Table \ref{tab:vla_results} presents results on the IndoorUAV-VLN task. Compared with the VLA setting, the challenges of long-range goal-directed planning amplify performance differences among models. Our IndoorUAV-Agent method achieves the best performance across all metrics, attaining 7.29\% and 5.06\% SR in seen and unseen environments respectively, while also achieving the highest NDTW (17.19\% / 15.65\%). Compared to the \(\bm{\pi_{0}}\) baseline, which uses the same low-level policy but without any task decomposition, our method improves SR by +4.37\% in the seen split and +2.23\% in the unseen split. These gains demonstrate the effectiveness of integrating instruction decomposition into the planning process.

While NaVid yields relatively high OSR scores (14.70\% / 16.21\%), its SR remains low (0.75\% / 0.84\%). Upon inspection, we find that NaVid often fails to predict the \texttt{stop} action. As a result, the agent tends to overshoot the goal region, leading to failures under the SR criterion, even when parts of the path are aligned with the instruction (thus yielding a higher OSR).

\subsection{Qualitative results}
Figure \ref{vis} presents several qualitative results of our IndoorUAV-Agent model on both VLA and VLN tasks.  
The top two examples illustrate short-range navigation scenarios, where the agent executes 2–3 actions to reach the target.  
The bottom example showcases a long-horizon VLN task. By decomposing the original instruction into a sequence of shorter sub-instructions, our model is able to successfully complete the long-range navigation.

\section{Conclusion}
We present IndoorUAV, the first large-scale benchmark for aerial VLN in indoor environments. Built with 1000+ environments, IndoorUAV features diverse indoor layouts and simulates UAV-specific viewpoints and motion dynamics, enabling research across both VLN and VLA paradigms. We further propose IndoorUAV-Agent, a modular framework designed to tackle the challenges of long-horizon aerial navigation through task decomposition and multi-granularity instruction understanding.

\section{Acknowledgements}
This work was supported by National Natural Science Foundation of China (Grant No.: 62372015), Key Laboratory of Intelligent Press Media Technology, and State Key Laboratory of General Artificial Intelligence.

\bibliography{CameraReady/LaTeX/aaai2026}

\clearpage
\lstset{%
	basicstyle={\footnotesize\ttfamily},
	numbers=left,numberstyle=\footnotesize,xleftmargin=2em,
	aboveskip=0pt,belowskip=0pt,
	showstringspaces=false,tabsize=2,breaklines=true}
\floatstyle{ruled}
\newfloat{listing}{tb}{lst}{}
\floatname{listing}{Listing}

\pdfinfo{
/TemplateVersion (2026.1)
}

\setcounter{secnumdepth}{0} 

 \title{ Supplementary Material}


\maketitle

\section{Implement Details}
For reproducibility, we provide detailed training configs and hyperparameter settings of the fine-tuned baseline models on the IndoorUAV Bench.

\begin{table}[h]
  \centering
    \centering
    \begin{tabular}{lc}
      \toprule
      \textbf{Parameter} & \textbf{Value} \\
      \midrule
      Batch Size & 32 \\
      Epochs & 10 \\
      Learning Rate & 1e-4 \\
      GPU & 1 × RTX 2080 Ti \\
      \bottomrule
    \end{tabular}
    \caption{Seq2Seq and CMA training config.}
    \label{tab:config1}
\end{table}

\begin{table}[h]
    \centering
    \begin{tabular}{lc}
      \toprule
      \textbf{Parameter} & \textbf{Value} \\
      \midrule
      Batch Size & 8 \\
      Max Learning Rate & 5e-4 \\
      LoRA & True \\
      LoRA Rank & 32 \\
      Max Training Steps & 300000\\
      GPU & 1 × A6000 \\
      \bottomrule
    \end{tabular}
    \caption{OpenVLA training config.}
    \label{tab:config2}
\end{table}

\begin{table}[h]
  \centering
    \centering
    \begin{tabular}{lc}
      \toprule
      \textbf{Parameter} & \textbf{Value} \\
      \midrule
      Batch Size & 32 \\
      Learning Rate & 5e-5 \\
      LoRA & True\\
      LoRA Rank & 32\\
      Action Horizon & 10\\
      Max Training Steps & 30000\\
      GPU & 2 × A6000 \\
      \bottomrule
    \end{tabular}
    \caption{\({\pi_{0}}\) and \({\pi_{0}}\)-FAST training config.}
    \label{tab:config3}
\end{table}

\begin{table}[h]
    \centering
    \begin{tabular}{lc}
      \toprule
      \textbf{Parameter} & \textbf{Value} \\
      \midrule
      Batch Size & 1 \\
      Learning Rate &  \\
      LoRA & True\\
      LoRA Rank & \\
      LoRA Alpha &  \\
      GPU & 4 × A6000 \\
      \bottomrule
    \end{tabular}
    \caption{Navid training config.}
    \label{tab:config4}
\end{table}

\begin{figure*}[ht]
\centering
\includegraphics[width=\textwidth]{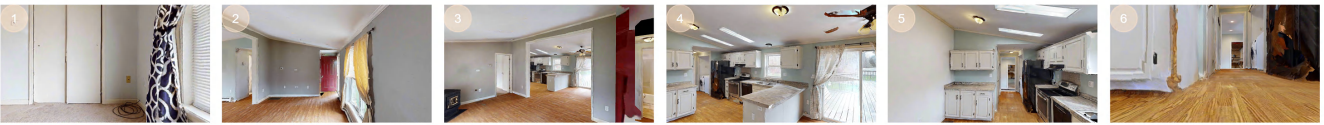}
\caption{Visualization of successful examples on both IndoorUAV-VLA and IndoorUAV-VLN.}
\label{vis_success}
\end{figure*}

\begin{figure*}[ht]
\centering
\includegraphics[width=\textwidth]{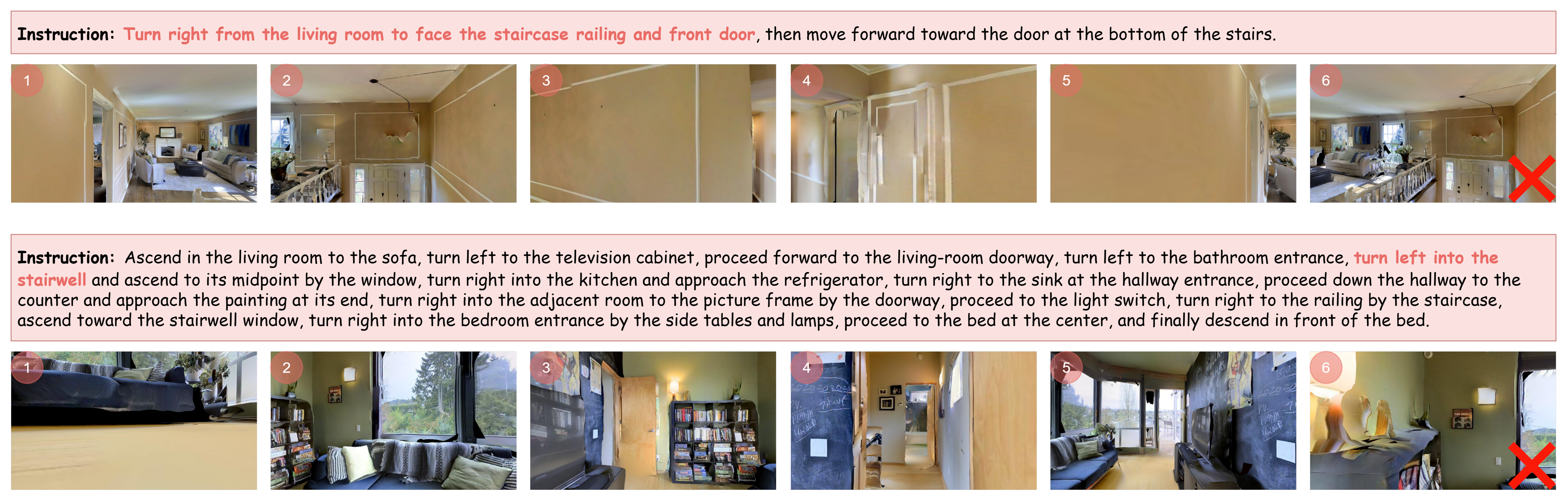}
\caption{Visualization of failed examples on both IndoorUAV-VLA and IndoorUAV-VLN. The IndoorUAV-Agent failed to successfully identify the target location of the current instruction, resulting in continuous performance of the current action without executing subsequent instructions. The sentences marked in red in the figure are the instructions that failed to execute.}
\label{vis_fail}
\end{figure*}
 
\section{Additional Qualitative Results}

 We provide more visualization results of our IndoorUAV-Agent on both IndoorUAV-VLA and IndoorUAV-VLN tasks here. Fig.~\ref{vis_success} and Fig.~\ref{vis_fail} present the successful and failed examples respectively.
 
\section{Prompt Template for Instruction Generation}

\begin{figure*}[t]
\centering
\includegraphics[width=0.98\textwidth]{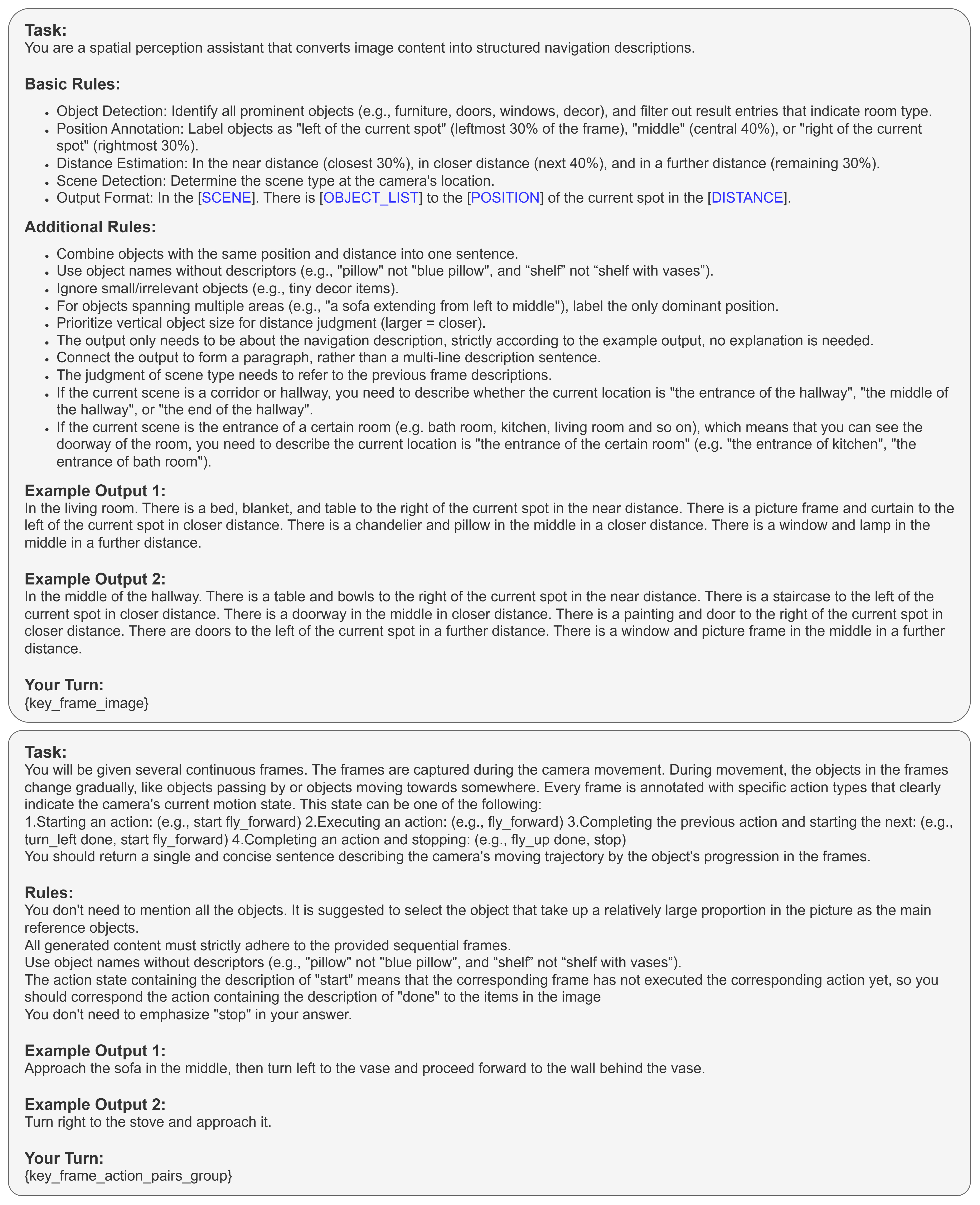}
\caption{Prompt for key frame captions generation (upper) and prompt for VLA instruction generation from key frame-action pairs group (lower).}
\label{prompt_1}
\end{figure*}

\begin{figure*}[t]
\centering
\includegraphics[width=0.97\textwidth]{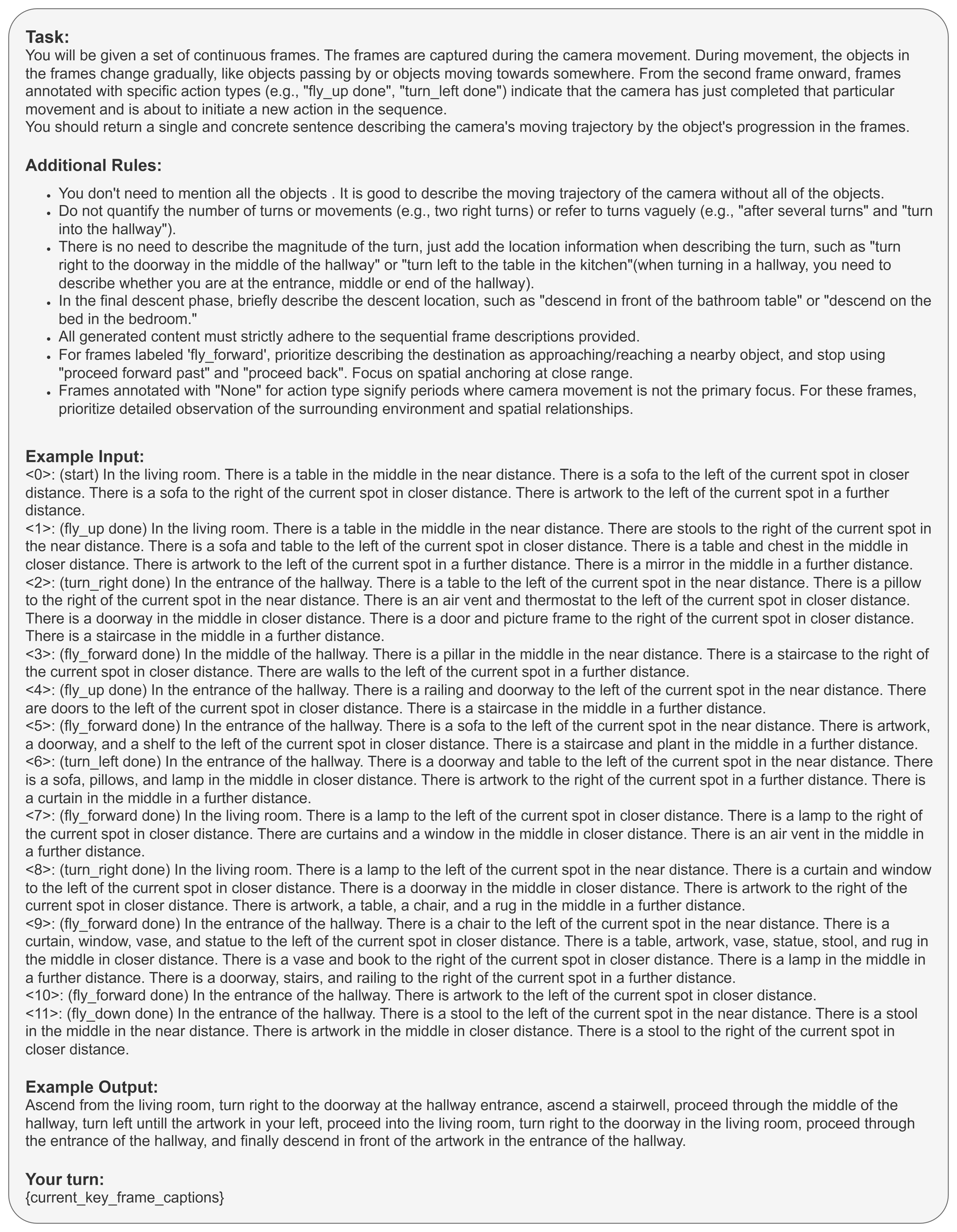}
\caption{Prompt for VLN instruction generation from key frame captions.}
\label{prompt_2}
\end{figure*}

As shown in Fig.~\ref{prompt_1} and .~\ref{prompt_2}, we use the three prompt templates to generate key frame captions, VLA instructions and VLN instructions respectively. 

For VLA tasks, due to their short-horizon characteristics, the correlations between their key frames are often quite obvious. So we choose to extract its key frame action state and form key frame-action pairs with the corresponding images, and then input the group of key frame-action pairs and the corresponding prompt template into ChatGPT to generate the corresponding VLA instructions.

However, for VLN tasks, due to their much longer trajectory path lengths, they often have far more key frames as well. Directly inputting so many key frame images to the LLM often leads to ambiguity and confusion in the instructions. Therefore, to ensure the generation of high-quality instructions, we first input the extracted key frame images and caption generation template to the LLM frame by frame to generate a sequence of captions for the key frames. Then, by combining the action state information, the caption sequence and the VLN instruction generation template are input to the ChatGPT to obtain the final VLN instruction.

\section{Evaluation Metric Calculation}

\subsection{Success Rate (SR)}

Due to the significant differences between VLA and VLN tasks, we adopt different success criteria for the two task types.

For VLA tasks, the predicted trajectory endpoint will be judged as successful only when the spatial distance from the target of the ground truth is less than 0.5m and the difference in direction angle is less than \(\frac{\pi}{4}\):
\begin{equation}
\text{Success}_{\text{VLA}} = 
\begin{cases}
1, & \text{if } 
\begin{aligned}
&\|\mathbf{p}_{\text{pred}} - \mathbf{p}_{\text{target}}\|_2 < 0.5 \\
&\text{and } |\theta_{\text{pred}} - \theta_{\text{target}}| < \frac{\pi}{4}
\end{aligned} \\
0, & \text{otherwise}
\end{cases}
\end{equation}

While for VLN tasks, a trajectory is judged as successful as long as the final position is within 2 meters of the target. 
\begin{equation}
\text{Success}_{\text{VLN}} = 
\begin{cases}
1, & \text{if } 
\|\mathbf{p}_{\text{pred}} - \mathbf{p}_{\text{target}}\|_2 < 2 \\
0, & \text{otherwise}
\end{cases}
\end{equation}

\subsection{Normalized Dynamic Time Warping (NDTW)}
The general formula for calculating the NDTW metric is:
\begin{equation}
  \mathrm{NDTW}(R, P) = \exp\left(-\frac{\mathrm{DTW}(R, P)}{L_R \cdot d_{\mathrm{th}}}\right)  
\end{equation}
where $R$ denotes the reference trajectory, $P$ denotes the predicted trajectory, $L_R$ represents the path length of the reference trajectory, and $d_{\mathrm{th}}$ indicates the distance threshold.

In previous work, the computation of NDTW typically considered only the 3D coordinate sequences of reference and predicted trajectories, disregarding the yaw angle variation sequences. However, in our IndoorUAV-VLA task, turning maneuvers constitute a significant portion of actions, and there exist numerous episodes composed solely of turning actions with zero path length. Consequently, for VLA tasks, relying solely on spatial coordinate matching is inadequate, and the alignment of yaw angle sequences must also be considered. 

To address this, we compute both the spatial NDTW ($\mathrm{NDTW}_\mathrm{s}$) for 3D coordinates and the rotational NDTW ($\mathrm{NDTW}_\mathrm{r}$) for yaw angles per episode. These components are adaptively weighted based on episode characteristics to form the final metric:

\begin{enumerate}
    \item Compute reference path length $L$ and cumulative rotation angle $\Theta$ per episode
    \item Normalize against dataset averages: 
    \[
    \widetilde{L} = \frac{L}{\overline{L}}, \quad 
    \widetilde{\Theta} = \frac{\Theta}{\overline{\Theta}}
    \]
    where $\overline{L} = 2.2 \text{ m}$ and $\overline{\Theta} = 1.0 \text{ rad}$ are the mean path length and rotation angle across all episodes
    \item Calculate the final metric as:
    \begin{equation}
    \mathrm{NDTW}_\mathrm{final} = 
    \frac{
        \widetilde{L} \cdot \mathrm{NDTW}_\mathrm{s} + 
        \widetilde{\Theta} \cdot \mathrm{NDTW}_\mathrm{r}
    }{
        \widetilde{L} + \widetilde{\Theta}
    }
    \end{equation}
\end{enumerate}

While for VLN tasks, we only calculate the spatial $\mathrm{NDTW}_\mathrm{s}$. And for distance threshold $d_{\mathrm{th}}$, we set $d_{\mathrm{th}} = 1$ for VLA tasks and $d_{\mathrm{th}} = 10$ for VLN tasks.

\end{document}